\documentclass{article}

\usepackage{PRIMEarxiv}

\usepackage[utf8]{inputenc} % allow utf-8 input
\usepackage[T1]{fontenc}    % use 8-bit T1 fonts
\usepackage{hyperref}       % hyperlinks
\usepackage{url}            % simple URL typesetting
\usepackage{booktabs}       % professional-quality tables
\usepackage{amsfonts}       % blackboard math symbols
\usepackage{nicefrac}       % compact symbols for 1/2, etc.
\usepackage{microtype}      % microtypography
\usepackage{lipsum}
\usepackage{fancyhdr}       % header
\usepackage{graphicx}       % graphics
\graphicspath{{media/}}     % organize your images and other figures under media/ folder
\usepackage{amssymb}
\usepackage{caption}
\usepackage{pifont}
\usepackage{multirow}
\usepackage{tabularx}
\usepackage{makecell}
\usepackage{array}
\usepackage{natbib}
\usepackage[table]{xcolor}
\usepackage{tikz}
\usepackage{xcolor}
\definecolor{seenblue}{RGB}{191,210,228}
\definecolor{unseenorange}{RGB}{240,212,198}
\definecolor{captionred}{RGB}{255,0,0}
\definecolor{captiongreen}{RGB}{0,170,90}

\definecolor{perblue}{RGB}{68,114,196}
\definecolor{perpurple}{RGB}{156,112,189}
\definecolor{chrxcolor}{RGB}{1, 140, 116}

\title{OpenEarth-OVS44: A Multimodal Remote Sensing Dataset for Open-Vocabulary Segmentation}
\author{
Yimin Wei$^{1,2}$ \quad Hongruixuan Chen$^{2}$ \quad Hao Zheng$^{2}$ \quad Hao Liu$^3$ \\
\textbf{Jian Song$^2$ \quad Yuting Zhu$^4$ \quad Junshi Xia$^2$ \quad Naoto Yokoya$^{1,2,\dagger}$}\\
$^1$The University of Tokyo \quad $^2$RIKEN AIP \quad $^3$East China Normal University \quad $^4$Tsinghua University \\
{\small $^{\dagger}$Corresponding author}
}

\begin{document}
\maketitle

\begin{abstract}
In remote sensing, open-vocabulary segmentation is particularly important for flexible and scalable land-cover understanding across diverse regions, as it improves generalization to novel categories without requiring exhaustive pixel-level annotations. Despite its great promise, progress in this area remains largely constrained by existing datasets with limited category coverage and overly coarse taxonomies, and therefore struggles to support fine-grained semantic understanding required in real-world applications such as urban functional zoning, carbon accounting, and post-disaster assessment. To address these challenges, we present {OpenEarth-OVS44}, a multimodal benchmark with 290,587 expert-verified segments from 2,832 RGB images and 1,341 strictly registered RGB-SAR pairs at sub-meter spatial resolution, covering 40 cities across France, Japan, and the United States under a unified 44-class taxonomy organized into 9 base categories. Three evaluation tracks are provided: an RGB-only fine-grained benchmark, a multimodal track with diverse synthetic cloudy conditions for weather-robust evaluation, and a held-out cross-region track. Extensive experiments on representative open-vocabulary segmentation models demonstrate that training on OpenEarth-OVS44 substantially improves fine-grained segmentation performance over the strongest training-free baseline, with improvements of up to 7.1 mIoU points on \textit{unseen} categories and 27.6 mIoU points on \textit{seen} categories. The dataset and code will be released.
\end{abstract}

% keywords can be removed

\section{Introduction}
\par Open-vocabulary segmentation (OVS) aims to assign semantic labels to image regions from an open-ended set of textual categories, enabling recognition beyond the classes observed during training. In remote sensing (RS), OVS is particularly valuable for flexible and scalable land-cover understanding across diverse geographical regions, without requiring exhaustive pixel-level annotations and fixed predefined class sets. Recently, OVS has achieved promising progress in unimodal remote sensing through the alignment of visual regions with arbitrary textual concepts~\citep{li2026exploring,ye2025towards}, and has further been extended to RGB-SAR fusion for handling cloud-degraded optical observations~\citep{wei2026mm}.

\par
\begin{figure*}[ht]
\centering
\includegraphics[width=\textwidth]{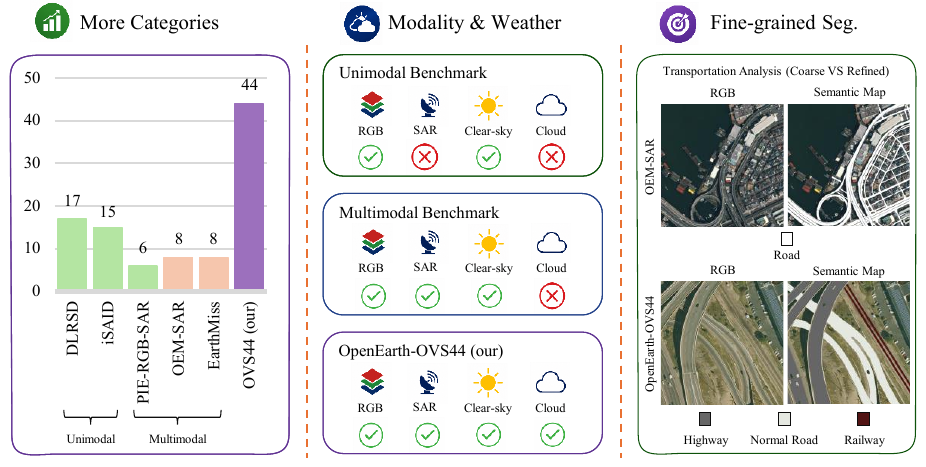}
\caption{Existing unimodal and multimodal benchmarks fail to support fine-grained segmentation due to limited categories and coarse taxonomies. In contrast, OpenEarth-OVS44, a multimodal and multi-weather benchmark, offers 44-class high-quality annotations, facilitating real-world RS applications. For example, in existing benchmarks, a single \textit{road} label may collapse functionally distinct categories such as \textit{highway}, \textit{normal road}, and \textit{railway}, whereas OpenEarth-OVS44 explicitly distinguishes them. DLRSD, iSAID, PIE, EarthMiss and OEM-SAR represent the DLRSD \citep{chaudhuri2017multilabel}, iSAID \citep{waqas2019isaid}, PIE-RGB-SAR \citep{ZHANG2024574}, EarthMiss \citep{zhou2026remote} and OpenEarthMap-SAR \citep{xia2025openearthmap} datasets, respectively.}
\label{fig:figcompare}
\end{figure*}

 Despite rapid progress, existing OVS studies in the RS domain~\citep{li2026exploring,wei2026mm} remain largely confined to current benchmarks with limited category coverage and overly coarse taxonomies. Specifically, unimodal RGB benchmarks~\citep{chaudhuri2017multilabel, waqas2019isaid} cap at 17 classes and multimodal RGB-SAR benchmarks~\citep{xia2025openearthmap, ZHANG2024574} at 8. Such coarse taxonomies collapse semantically and functionally distinct categories, limiting fine-grained semantic understanding in real-world remote sensing applications (Figure~\ref{fig:figcompare}). As further demonstrated in Table~\ref{tab:taxonomy_granularity}, strong performance under coarse taxonomies can conceal substantial fine-grained recognition failures. This limitation is particularly critical for applications that require distinguishing functionally distinct yet visually similar categories. Furthermore, real-world RS observations are frequently affected by cloud or haze contamination, making multimodal RGB-SAR fusion essential for robust and generalizable scene understanding. Consequently, the lack of fine-grained multimodal remote sensing benchmarks has become a major bottleneck for progress in RS OVS.

% Such coarse taxonomies collapse semantically and functionally distinct categories, failing to support the fine-grained semantic understanding required in real-world remote sensing applications (Figure~\ref{fig:figcompare}). This limitation is particularly critical because accurate interpretation in RS scenarios, such as urban functional zoning, carbon accounting, and post-disaster assessment, depends on distinguishing functionally distinct yet visually similar categories. Furthermore, real-world RS observations are frequently affected by cloud or haze contamination, making multimodal RGB-SAR fusion essential for robust and generalizable scene understanding. Consequently, the lack of fine-grained multimodal remote sensing benchmarks has become a major bottleneck for progress in RS OVS.

% Such coarse taxonomies collapse semantically and functionally distinct categories, thereby masking the fine-grained limitations of existing OVS models. As further demonstrated in Table~\ref{tab:taxonomy_granularity}, strong performance under coarse taxonomies can conceal substantial failures in fine-grained semantic discrimination. This limitation is particularly critical for real-world remote sensing applications, where accurate interpretation often requires distinguishing functionally distinct yet visually similar categories.

\par Closing this gap is considerably more challenging than simply scaling existing annotation pipelines. Fine-grained remote sensing annotation differs from natural-image annotation in three ways that constrain the scalability of such benchmarks. First, the visual signatures separating fine-grained classes are subtle and context-dependent. Greenhouses are recognized by translucent roofs, whereas shopping malls and factories are distinguished mainly through surrounding cues such as parking footprint and loading-dock layout, requiring remote sensing interpretation training that crowdsourced annotators do not possess. Second, RGB-SAR annotation requires expert cross-validation across modalities with different viewing geometry and non-photographic SAR appearance. Third, fine-grained taxonomies amplify inter-annotator ambiguity at boundary cases (e.g., \textit{residential building} vs. \textit{shopping mall}, \textit{normal road} vs. \textit{highway}), demanding adjudication rather than majority vote. Therefore, introducing a fine-grained multimodal RS benchmark remains an open challenge.

% However, constructing such a benchmark is profoundly challenging, presenting three key obstacles. Firstly, fine-grained annotation in RS requires substantial domain expertise to distinguish subtle visual patterns and infer functional categories from contextual cues. For instance, greenhouses feature transparent or translucent structures, whereas functional categories such as shopping malls and factories are more reliably identified from their surrounding environments, such as large parking areas or dense industrial facilities. Secondly, compared with coarse labeling, annotating a large set of fine-grained categories is considerably more labor-intensive and time-consuming. Thirdly, annotation ambiguity across annotators can further introduce label noise, making expert review essential for ensuring label accuracy and consistency. Consequently, introducing a fine-grained multimodal RS benchmark remains an open challenge.

\par
\begin{figure*}[!t]
\centering
\includegraphics[width=\textwidth]{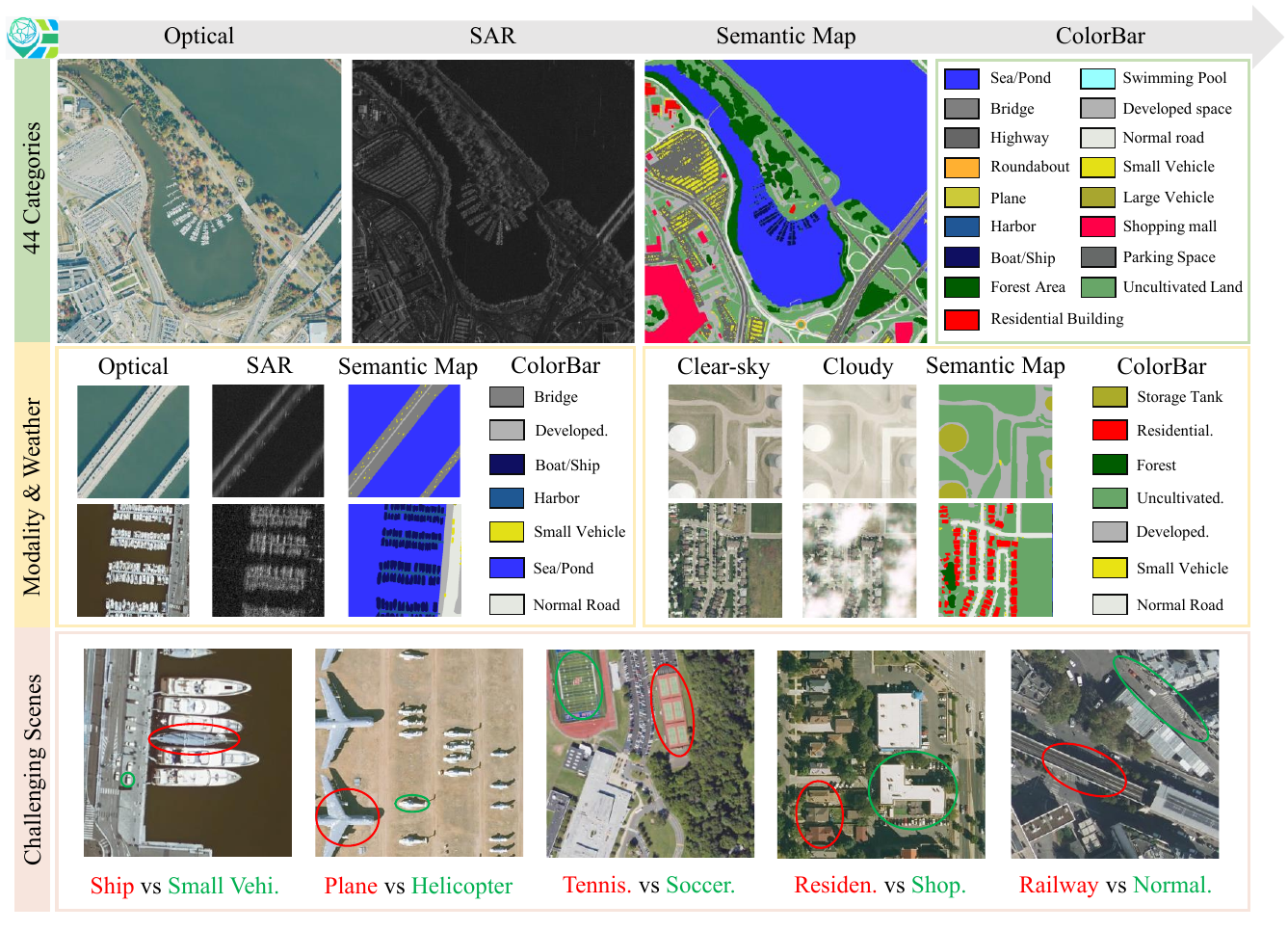}
\caption{\textbf{Overview of {OpenEarth-OVS44}.} Our benchmark exhibits three key properties: (1) broad category coverage with 44 fine-grained manually annotated classes; (2) multiple remote sensing modalities and weather conditions, including RGB and SAR imagery under clear-sky and synthetic cloudy conditions; and (3) highly challenging scenes in which visually similar fine-grained categories frequently coexist, such as \textit{ship} vs. \textit{small vehicle}, \textit{plane} vs. \textit{helicopter}, \textit{tennis court} vs. \textit{soccer field}, \textit{residential building} vs. \textit{shopping mall}, and \textit{railway} vs. \textit{normal road}. The co-occurring categories are highlighted by \textcolor{captionred}{red} and \textcolor{captiongreen}{green} circles, respectively.}
\label{fig:fig1}
\end{figure*}

 Within these constraints, we introduce {OpenEarth-OVS44}, a fine-grained multimodal remote sensing benchmark designed for both unimodal and multimodal open-vocabulary segmentation. {OpenEarth-OVS44} comprises 290,587 segments annotated from 2,832 RGB images of size 512 × 512, covering 40 cities across France, Japan and the United States. It further includes 1,341 paired RGB-SAR images at sub-meter spatial resolution and manual annotations spanning 44 classes. As illustrated in Figure \ref{fig:fig1}, {OpenEarth-OVS44} is characterized by three key properties. (1) \textit{Broad category coverage with fine-grained annotations}. It provides, to the best of our knowledge, the largest category set among existing remote sensing OVS datasets, comprising 44 fine-grained classes organized into 9 base categories. (2) \textit{Multiple modalities and weather conditions}. The benchmark includes two sensing modalities, RGB and SAR, and is organized into three evaluation tracks. The first track is an RGB-only OVS benchmark comprising 1,936 RGB-semantic map pairs annotated with 44 fine-grained categories. The second is a multimodal track containing 1,341 strictly registered RGB-SAR-semantic map triplets with 43 shared categories. Its RGB images cover diverse synthetic cloudy conditions, including thin and thick cloud cover, enabling systematic evaluation of multimodal fusion under severely degraded optical observations. All images in these two tracks are sourced from France and the United States. The third is a held-out cross-region track comprising 896 RGB-semantic map pairs acquired over Japan and annotated with 40 fine-grained categories, and is used entirely for cross-domain testing.  (3) \textit{Highly challenging remote sensing scenes}. The benchmark is intentionally designed to be challenging, as multiple fine-grained categories frequently coexist within a single image.

Through comprehensive benchmarking experiments on {OpenEarth-OVS44}, we observe that cutting-edge training-free RS OVS models struggle with fine-grained RS OVS tasks. In addition, five state-of-the-art OVS models are trained on our dataset, establishing strong baselines across all evaluation tracks. The main contributions of this paper are summarized as follows:

1. To advance both unimodal and multimodal OVS, we propose {OpenEarth-OVS44}, a multi-weather and multi-sensor remote sensing benchmark with three evaluation tracks. It contains 290,587 manually annotated segments covering 44 classes, derived from 2,832 RGB images and 1,341 paired RGB-SAR images at sub-meter spatial resolution.

2. We systematically evaluate the fine-grained OVS capability of existing models, including two advanced training-free RS OVS methods, four unimodal OVS models and a multimodal OVS framework on {OpenEarth-OVS44}. The comparative analysis reveals their strengths and limitations, while highlighting several critical directions for future progress in RS open-vocabulary segmentation.

3. We establish strong baselines by training five representative OVS models, spanning both natural-image and remote-sensing domains, on {OpenEarth-OVS44}. The top-performing trained models outperform the strongest training-free baseline by
up to 7.1 mIoU points on \textit{unseen} categories and 27.6 mIoU points on \textit{seen} categories, highlighting the value of OpenEarth-OVS44 as a benchmark for fine-grained remote sensing OVS.

\section{Related Work}

% \subsection{}
\textbf{Open-Vocabulary Segmentation Model.} Leveraging large-scale vision-language models (VLMs), including CLIP \citep{radford2021learning} and ALIGN \citep{jia2021scaling}, OVS models \citep{li2022language, ding2022decoupling, xu2022simple, ding2022open, yu2023convolutions, xu2023side, xie2024sed} associate arbitrary textual concepts with visual regions, thereby enabling recognition beyond fixed training categories. Representative OVS methods \citep{zhou2022extract, ghiasi2022scaling, liang2023open, cho2024catseg, shan2024open, li2025fgaseg} have achieved remarkable performance on natural-image benchmarks. However, these methods are primarily developed for natural images and do not explicitly address the unique challenges posed by remote sensing imagery, such as large scale variation, arbitrary object orientations, and complex spatial distributions.

\par \textbf{Open-Vocabulary Segmentation Model in RS. }OVS has recently been extended from natural to RS imagery, where CLIP-based methods are typically grouped into training-free and training-required approaches. Training-free methods adapt CLIP to RS with minimal architectural change, exemplified by SegEarth-OV~\citep{li2025segearth} and its SAM3-based extension SegEarth-OV3~\citep{li2025segearth3} . Training-required methods learn from labeled base classes for better domain adaptation and local detail, including rotation-aware~\citep{cao2024open} and DINO-augmented~\citep{ye2025towards, li2026exploring} designs. All of these remain unimodal and degrade under low-visibility conditions. To address this, MM-OVSeg~\citep{wei2026mm} introduces optical-SAR fusion \citep{li2022mcanet,wu2023crofusenet} that aligns complementary RGB and SAR features with text, improving robustness under cloudy conditions. Despite this progress, all current RS OVS methods are evaluated on coarse-taxonomy benchmarks, leaving fine-grained capability essentially unmeasured.

\par \textbf{Open-Vocabulary Segmentation Benchmarks. }OVS has achieved substantial success in the natural image domain, largely driven by the availability of large-scale semantic segmentation datasets \citep{zhou2019semantic, everingham2010pascal, mottaghi2014role} with rich semantic categories, such as COCO-Stuff \citep{caesar2018coco} with 171 classes. In remote sensing, existing OVS benchmarks are typically derived by relabeling mainstream RS semantic segmentation datasets~\citep{wang2021loveda,xia2023openearthmap,lyu2020uavid,cai2025vdd, chen2018large, ji2018fully, maggiori2017can, gupta2019xbd, zhu2021global, mnih2013machine, van2018spacenet} under a seen/unseen split. They fall into two regimes: unimodal RGB benchmarks~\citep{chaudhuri2017multilabel, waqas2019isaid} with up to 17 classes, and multimodal RGB-SAR benchmarks~\citep{xia2025openearthmap, ZHANG2024574} with up to 8 classes. Two limitations persist across both. First, taxonomies are coarse: functionally distinct entities are merged into a single label, preventing fine-grained capability from being measured. Second, multimodal coverage is shallow: the multimodal regime is restricted to a handful of categories, leaving weather-robust fine-grained OVS without a benchmark. OpenEarth-OVS44 addresses both by providing 44 manually verified categories with strictly registered RGB-SAR pairs, enabling fine-grained evaluation in unimodal and multimodal settings under unified protocols.

%%%%%%%%%%%%%%%%%%%%% Section 3 %%%%%%%%%%%%%%%%%%%%%%%%%%%%%%
\section{OpenEarth-OVS44 Benchmark}

\subsection{Overview}

\par OpenEarth-OVS44 is designed to evaluate fine-grained OVS in remote sensing under three practical challenges: refined semantic granularity, optically degraded observations, and geographic domain shift. Unlike existing RS OVS benchmarks derived mainly from coarse semantic segmentation datasets, it provides expert-verified annotations under a unified 44-class taxonomy and supports both RGB and multimodal RGB-SAR evaluation. The benchmark contains 2,832 RGB images of size $512 \times 512$, among which 1,341 images have strictly registered SAR counterparts. In total, it provides 290,587 annotated segments from 40 cities across France, Japan, and the United States. The 44 categories are hierarchically organized into 9 base categories. OpenEarth-OVS44 consists of three evaluation settings. OpenEarth-OVS44-Optical evaluates fine-grained RGB-only OVS on France and US imagery. OpenEarth-OVS44-Cloudy evaluates RGB-SAR OVS under synthetic thin- and thick-cloud degradation. OpenEarth-OVS44-Cross uses Japan imagery as a held-out geographic domain to evaluate cross-region generalization.

\subsection{Benchmark Construction}

\subsubsection{Data Collection}

We construct {OpenEarth-OVS44} using multimodal remote sensing data, including optical and SAR imagery, collected from 40 cities across France, Japan and the United States. The SAR data are acquired from Umbra Space in Spotlight mode, while the optical data are aerial images sourced from the National Agriculture Imagery Program (NAIP) \footnote{\url{https://datagateway.nrcs.usda.gov}}, the Geospatial Information Authority of Japan (GSI) \footnote{\url{https://maps.gsi.go.jp/development/ichiran.html}}, and the French National Institute of Geographic and Forest Information (IGN) \footnote{\url{https://geoservices.ign.fr/bdortho}}. The optical imagery has a ground sampling distance of 0.15--0.5 m, while the SAR imagery has a spatial resolution of 0.2--0.6 m. The optical imagery contains RGB bands, whereas the SAR imagery mainly provides amplitude information in the VV or HH polarization. After quantization and 8-bit standardization, residual misalignment between the two geocoded modalities is further corrected through expert manual registration and cross-validation. Registration quality is further verified quantitatively, yielding an average RGB-SAR registration RMSE of 1.369 pixels, comparable to that of an established registered RGB-SAR benchmark \citep{Chen2025Bright}. To ensure a highly challenging benchmark, all samples were selected through a rigorous screening process conducted by an expert team of six annotators and five PhD researchers. A central criterion was that each image should contain multiple fine-grained categories within the same coarse category. Following this protocol, the dataset comprises 2,832 RGB images and 1,341 paired RGB-SAR images.

\subsubsection{Annotation and Quality Control}

We construct the taxonomy hierarchically by first defining nine base categories (as illustrated in Table \ref{tab:category_hierarchy}) based on commonly used category systems in existing remote-sensing datasets \citep{waqas2019isaid, xia2025openearthmap}. We then expand them into 44 refined categories according to differences in land-cover type and functional use, aiming to cover representative categories in practical remote-sensing applications. For the seen/unseen split, following MM-OVSeg~\citep{wei2026mm}, all nine refined classes under the \textit{Road} and \textit{Water} base categories are held out as \textit{unseen} to maintain comparability with prior RS OVS evaluations. To broaden \textit{unseen}-category coverage, we further include four practically meaningful but underrepresented categories---\textit{windmill}, \textit{beach}, \textit{barren/exposed soil}, and \textit{expressway toll station}---which are rarely treated as independent classes in existing RS OVS datasets. Including them in the \textit{unseen} split enables evaluation of model generalization to underrepresented concepts that are not covered by conventional coarse taxonomies. Their annotations are excluded from training, ensuring that no direct class-level supervision is provided. The final split contains 31 \textit{seen} and 13 \textit{unseen} classes (as listed in Table~\ref{tab:seenUnseen}), approximately a 7:3 ratio, comparable to commonly adopted OVS protocols~\citep{wei2026mm}. \textit{Seen} and \textit{unseen} categories account for 79.5\% and 20.5\% of annotated pixels, respectively.

Based on this taxonomy, annotators labeled all RGB images according to unified category definitions. To mitigate label noise caused by inter-annotator ambiguity, all initial annotations in {OpenEarth-OVS44} underwent a rigorous expert review process requiring approximately 512 person-hours. Three PhD researchers reviewed the initial annotations to ensure their accuracy and consistency, while two senior experts performed a cross-check and provided the final adjudication. This procedure ultimately produced 44-class high-quality annotations with 290,587 expert-verified segments.

%-----------------------------------------
\newcolumntype{Y}{>{\raggedright\arraybackslash}X}

% Small color swatch
\newcommand{\swatch}[3]{%
  \tikz[baseline=-0.1ex]\filldraw[fill={rgb,255:red,#1;green,#2;blue,#3},draw=black!40,line width=0.2pt]
  (0,0) rectangle (0.20,0.20);
}

% Swatch + category name
\newcommand{\cat}[4]{\swatch{#1}{#2}{#3}\,#4}
%-----------------------------------------
\begin{table*}[t]
\centering
\caption{Base categories and their corresponding refined categories in {OpenEarth-OVS44}. Color swatches indicate the annotation color of each refined category.}
\label{tab:category_hierarchy}
\footnotesize
\setlength{\tabcolsep}{4pt}
\renewcommand{\arraystretch}{1.15}
\begin{tabularx}{\textwidth}{p{0.18\textwidth} >{\centering\arraybackslash}p{0.08\textwidth} Y}
\toprule
\textbf{Base Category} & \textbf{\#Refined} & \textbf{Refined Categories} \\
\midrule
Building 
& 6
& \cat{255}{227}{102}{greenhouse}, 
\cat{31}{112}{54}{stadium}, 
\cat{255}{0}{72}{shopping mall}, 
\cat{255}{51}{0}{factory}, 
\cat{163}{0}{0}{terminal}, 
\cat{255}{0}{0}{residential building} \\

Sports Field 
& 6
& \cat{144}{238}{144}{ground track field}, 
\cat{16}{143}{16}{soccer field}, 
\cat{23}{194}{117}{baseball diamond}, 
\cat{214}{108}{0}{basketball court}, 
\cat{255}{215}{0}{tennis court}, 
\cat{51}{255}{51}{golf course} \\

Transport Facilities 
& 5
& \cat{76}{76}{76}{train station}, 
\cat{205}{205}{205}{airport}, 
\cat{102}{105}{104}{parking space}, 
\cat{32}{87}{150}{harbor}, 
\cat{130}{126}{126}{expressway toll station} \\

Industrial Facilities 
& 6
& \cat{245}{245}{0}{tower}, 
\cat{153}{204}{255}{windmill}, 
\cat{92}{46}{0}{chimney}, 
\cat{67}{69}{71}{container crane}, 
\cat{0}{147}{255}{container}, 
\cat{171}{171}{41}{storage tank} \\

Road 
& 6
& \cat{255}{177}{51}{roundabout}, 
\cat{82}{20}{20}{railway}, 
\cat{127}{127}{127}{bridge}, 
\cat{102}{102}{102}{highway}, 
\cat{229}{232}{225}{normal road}, 
\cat{153}{153}{153}{airport runway} \\

Transport Vehicles 
& 6
& \cat{29}{87}{145}{train}, 
\cat{15}{15}{97}{boat/ship}, 
\cat{163}{163}{0}{helicopter}, 
\cat{204}{202}{57}{plane}, 
\cat{168}{166}{47}{large vehicle}, 
\cat{227}{224}{23}{small vehicle} \\

Water 
& 3
& \cat{51}{51}{255}{sea/lake/pond}, 
\cat{0}{0}{163}{river}, 
\cat{153}{255}{255}{swimming pool} \\

Vegetation 
& 3
& \cat{0}{92}{0}{forest}, 
\cat{0}{214}{0}{agricultural land}, 
\cat{104}{166}{104}{uncultivated land} \\

Non-vegetation 
& 3
& \cat{255}{153}{51}{beach}, 
\cat{135}{76}{27}{barren/exposed soil}, 
\cat{178}{178}{178}{developed space} \\
\bottomrule
\end{tabularx}
\end{table*}
%-----------------------------------------
% \newcolumntype{Y}{>{\centering\arraybackslash}X}

% \begin{table*}[h!]
% \centering
% \caption{Base categories and their corresponding refined categories in OpenEarth-OVS44.}
% \label{tab:category_hierarchy}
% \small
% % \setlength{\tabcolsep}{4.5pt}
% \renewcommand{\arraystretch}{1.18}
% % \resizebox{\linewidth}{!}{
% \begin{tabularx}{\textwidth}{>{\centering\arraybackslash}p{0.23\textwidth} Y}
% \toprule
% \textbf{Base Category} & \textbf{Refined Category} \\
% \midrule
% Building 
% & greenhouse, stadium, shopping mall, factory, terminal, residential building \\

% Sports Field 
% & ground track field, soccer field, baseball diamond, basketball court, tennis court, golf course \\

% Transportation Facilities 
% & train station, airport, parking space, harbor, expressway toll station \\

% Industrial Facilities 
% & tower, windmill, chimney, container crane, container, storage tank \\

% Road 
% & roundabout, railway, bridge, highway, normal road, airport runway \\

% Transportation Vehicles 
% & train, boat/ship, helicopter, plane, large vehicle, small vehicle \\

% Water 
% & sea/lake/pond, river, swimming pool \\

% Vegetation 
% & forest, agricultural land, uncultivated land \\

% Non-vegetation 
% & beach, barren/exposed soil, developed space \\
% \bottomrule
% \end{tabularx} %}
% \end{table*}

%-----------------------------------------

\subsubsection{Data Split}\label{sec:323}

% The dataset is organized into three evaluation settings, each with dedicated training, validation, and test splits.
The dataset includes three evaluation settings: OpenEarth-OVS44-Optical and OpenEarth-OVS44-Cloudy use training, validation, and test splits, while OpenEarth-OVS44-Cross is reserved for testing.

% \noindent {1).} {OpenEarth-OVS44-Optical} is a RGB-only OVS benchmark containing 1,936 RGB-semantic map pairs annotated with 44 fine-grained categories, with 1,008/160/768 images for training, validation, and testing, respectively. Following the standard OVS protocol, the 44 categories are divided into 31 seen and 13 unseen classes. The detailed seen/unseen split is provided in Table \ref{tab:seenUnseen}. All images in this benchmark are sourced from France and the United States.

\noindent {1).} {OpenEarth-OVS44-Optical} evaluates RGB-only fine-grained OVS on France and US imagery. It contains 1,936 RGB-semantic map pairs with 44 categories, split into 1,008 training, 160 validation, and 768 test images. The 44 categories are divided into 31 \textit{seen} and 13 \textit{unseen} classes.

% %----------------------------------------
% \begin{table*}[htbp]
% \renewcommand{\arraystretch}{1.35}
% \centering
% \caption{\label{tab:seenUnssen}{Detailed seen/unseen split in{OpenEarth-OVS44-Optical} benchmark.}}
% \resizebox{\linewidth}{!}{
% \begin{tabular}{|c|c|}
% \hline
% \textbf{Category Type} & \textbf{Category Name} \\ \hline
% % \midrule
% Seen & 
% \begin{tabular}{@{}c@{}}
% train station, airport, parking space, harbor, tower, chimney, container crane, container, storage tank, agricultural land, uncultivated land  \\
% forest, developed space, ground track field, soccer field, baseball diamond, basketball court, golf course, tennis court, train, boat/ship  \\
% helicopter, plane, large vehicle, small vehicle, greenhouse, residential building, terminal, factory, stadium, shopping mall   
% \end{tabular} \\  \hline
% % \midrule
% Unseen  & 
% \begin{tabular}{@{}c@{}}
% roundabout, railway, bridge, highway, normal road, airport runway, sea/lake/pond, river, swimming pool, windmill, beach \\ barren/exposed soil, expressway toll station 
% \end{tabular} \\\hline
% % \bottomrule
% \end{tabular}}
% \end{table*}

\begin{table*}[t]
\centering
\caption{Detailed seen/unseen category division in the {OpenEarth-OVS44-Optical} benchmark, including 31 \textit{seen} categories and 13 \textit{unseen} categories.}
\label{tab:seenUnseen}
\small
\setlength{\tabcolsep}{4pt}
\renewcommand{\arraystretch}{1.15}
\begin{tabularx}{\textwidth}{p{0.06\textwidth} >{\centering\arraybackslash}p{0.08\textwidth} >{\raggedright\arraybackslash}X}
\toprule
\textbf{Type} & \textbf{Num.} & \textbf{Category} \\
\midrule
Seen & 31 & train station, airport, parking space, harbor, tower, chimney, container crane, container, storage tank, agricultural land, uncultivated land, forest, developed space, ground track field, soccer field, baseball diamond, basketball court, golf course, tennis court, train, boat/ship, helicopter, plane, large vehicle, small vehicle, greenhouse, residential building, terminal, factory, stadium, shopping mall \\ \hline
Unseen & 13 & roundabout, railway, bridge, highway, normal road, airport runway, sea/lake/pond, river, swimming pool, windmill, beach, barren/exposed soil, expressway toll station \\
\bottomrule
\end{tabularx}
\end{table*}
%--------------------------------
%---------------0.9\textwidth--------------------------
% \begin{figure*}[!t]
% \centering
% \includegraphics[width=\textwidth]{Statistic.pdf}
% \caption{{The statistics of semantic annotations in {OpenEarth-OVS44} benchmark. \textcolor{seenblue}{Blue} bars and \textcolor{unseenorange}{orange} bars represent \textit{seen} and \textit{unseen} classes, respectively.}}
% \label{fig:statistic}
% \end{figure*}
%-----------------------------------------
%-----------------------------------------
% in terms of Sample Numbers, Resolutions, Category Numbers, and Segmentation Mask Availability

\begin{table*}[h!]
\renewcommand{\arraystretch}{1.5}
\centering
\caption{\label{tab:compareOther}Comparison with existing datasets used to construct remote sensing open-vocabulary segmentation benchmarks.}
\resizebox{\linewidth}{!}{
\begin{tabular}{l|c|c|c|c|c|c|c|c}
\hline
\textbf{Dataset}     & \textbf{Proposed Time} & \textbf{Image Width} & \textbf{\# Instances} & \textbf{\# Samples} & \textbf{RGB Res. (m)} & \textbf{SAR Res. (m)} & \textbf{\# Categories} & \textbf{Paired SAR Image} \\ \hline
DLRSD \citep{chaudhuri2017multilabel}  & 2018      &  256   &   184,033    & 2,100     &   0.3    &  --         & 17 & \ding{55} \\ %\hline
iSAID \citep{waqas2019isaid}  & 2019     &  800 $\sim$ 13,000    &  655,451    & 2,806            &  --    &   --  & 15 & \ding{55} \\ %\hline
DDHR-SK \citep{ren2022dual} & 2022     &   256   & 76,002     & 6,173      &    1   &   1       & 5 & \checkmark \\ %\hline
DDHR-CH \citep{ren2022dual} & 2022     &   256   &  25,307    & 2,122      &    1   &   1       & 5 & \checkmark \\ %\hline
PIE-RGB-SAR \citep{ZHANG2024574} & 2024     &   256   & 43,288     & 4,865      &   0.5   &   3       & 6 & \checkmark \\ %\hline
% BRIGHT \citep{Chen2025Bright} & 2025     &  1,024   &      & 4,246            & 0.3 $\sim$ 1    &     & 7 & \checkmark \\ \hline
OpenEarthMap-SAR \citep{xia2025openearthmap}              & 2025      &   1,024    &  1,477,625   & 5,033            & 0.15 $\sim$ 0.5   &   0.2 $\sim$ 0.6   & 8 &  \checkmark \\ %\hline
% EarthMiss \citep{zhou2026remote}          & 2026      &   1,024   &     & 3,355           & 0.6    &  0.35   & 8 &   \checkmark                   \\ \hline
OpenEarth-OVS44 (Ours)         & 2026   &  512  & 290,587 & 2,832  & 0.15 $\sim$ 0.5  &     0.2 $\sim$ 0.6   & \textbf{44} &  \checkmark       \\ \hline
\end{tabular}}
\end{table*}

%---------------------------------------
% ``expressway toll station''

\noindent {2).} {OpenEarth-OVS44-Cloudy} 
evaluates multimodal RGB-SAR OVS under optically degraded conditions. It contains 1,341 registered RGB-SAR-semantic map triplets, split into 641 training, 123 validation, and 577 test samples. Since \textit{expressway toll station} appears only in the RGB modality, it is excluded from the multimodal shared-category evaluation. The final evaluation includes 43 shared categories, with 31 \textit{seen} and 12 \textit{unseen} classes. Following the protocol of MM-OVSeg~\citep{wei2026mm}, we use SatelliteCloudGenerator~\citep{rs15174138} to synthesize mixed cloud contamination for each RGB image, resulting in a cloudy benchmark that covers both thick- and thin-cloud conditions. The corresponding SAR image is kept unchanged across these RGB variants.

% , for each RGB image, we use SatelliteCloudGenerator~\citep{rs15174138} to synthesize mixed cloud contamination, resulting in a cloudy benchmark that covers both thick- and thin-cloud conditions. 

% contains 1,341 strictly registered RGB-SAR-semantic map triplets for multimodal OVS, split into 641/123/577 training, validation, and test samples. Notably, RGB modality and SAR modality share 43 categories, while ‘expressway toll station’ appears only in the RGB modality. After excluding this category from the unseen set, this benchmark finally contains 31 seen categories and 12 unseen categories. This benchmark is designed to evaluate the robustness of multimodal OVS models under cloudy conditions. Specifically, we apply SatelliteCloudGenerator \citep{rs15174138} to the RGB images to synthesize both thick and thin cloud cover. 

\noindent {3).} {OpenEarth-OVS44-Cross} 
evaluates cross-region generalization. It contains 896 RGB-semantic map pairs acquired over Japan and is used entirely for testing. No Japan image is used for training or validation. Four categories, including \textit{helicopter}, \textit{plane}, \textit{airport runway}, and \textit{windmill}, are absent from this subset, resulting in 40 evaluated categories with 29 \textit{seen} and 11 \textit{unseen} classes.

% comprises 896 RGB-semantic map pairs acquired over Japan and annotated with 40 fine-grained categories, and is used entirely for cross-domain evaluation. Notably, four categories—helicopter, plane, airport runway, and windmill—are absent from this benchmark, resulting in a final split of 29 seen categories and 11 unseen categories.

% \noindent {2).} OpenEarth-OVS44-Clear contains 1,341 strictly registered RGB-SAR-semantic map triplets for multimodal OVS, split into 641/123/577 training, validation, and test samples. Notably, RGB modality and SAR modality share 43 categories, while ‘expressway toll station’ appears only in the RGB modality. After excluding these two categories from the unseen set, this benchmark finally contains 31 seen categories and 12 unseen categories.
% \noindent {3).}{OpenEarth-OVS44-Optical} is designed to evaluate the robustness of multimodal OVS models under cloudy conditions. Specifically, we apply SatelliteCloudGenerator \citep{rs15174138} to the RGB images to synthesize both thick and thin cloud cover. Its data split and seen/unseen category partition follow OpenEarth-OVS44-Clear.

% \subsection{OpenEarth-OVS44 Statistics}

\subsection{Dataset Statistics and Comparison}

% Table~\ref{tab:compareOther} compares OpenEarth-OVS44 with existing RS OVS-related datasets. Compared with prior RGB benchmarks, OpenEarth-OVS44 provides substantially richer category granularity. Compared with existing RGB-SAR benchmarks, it provides a much larger fine-grained taxonomy while retaining paired multimodal observations. These properties make OpenEarth-OVS44 suitable for evaluating not only RGB-only OVS, but also weather-robust multimodal OVS and cross-region generalization.

% %%%%% Delete Figure 3%%%%%%%%%%%%%%%
% Additionally, our {OpenEarth-OVS44} dataset comprises a total of 44 categories. These categories cover diverse representative remote sensing scenarios, including urban areas, rural areas, airports and harbors. In addition to common categories present in existing datasets, {OpenEarth-OVS44} also includes a large number of rare but practically important categories, such as \textit{windmill}, \textit{container crane}, and \textit{expressway toll station}. Furthermore, unlike existing datasets that treat \textit{sports field} as a single coarse category, our benchmark decomposes it into six fine-grained classes: \textit{ground track field}, \textit{soccer field}, \textit{baseball diamond}, \textit{basketball court}, \textit{tennis court}, and \textit{golf course}. This refined category design provides a stronger basis for evaluating fine-grained OVS in real-world RS applications.

Table~\ref{tab:compareOther} compares OpenEarth-OVS44 with existing RS OVS-related datasets. OpenEarth-OVS44 provides substantially finer semantic granularity than prior RGB benchmarks while retaining paired RGB-SAR observations, enabling unified evaluation of RGB-only, weather-robust multimodal, and cross-region OVS. Its 44-class taxonomy covers diverse remote-sensing scenarios and includes both common categories and underrepresented but practically important concepts, such as \textit{windmill}, \textit{container crane}, and \textit{expressway toll station}. This fine-grained design provides a stronger basis for evaluating semantic discrimination in real-world RS applications.

%%%%%%%%%%%%%%%%%%%%% Section 4 %%%%%%%%%%%%%%%%%%%%%%%%%%%%%%
\section{Benchmark Experiments}
\subsection{Baselines and Experimental Setups}\label{sec:baselines}

We evaluate representative OVS models from four categories: natural-image OVS methods, including CAT-Seg~\citep{cho2024catseg} and EBSeg~\citep{shan2024open}; unimodal remote sensing OVS methods, including GSNet~\citep{ye2025towards} and RSKT-Seg~\citep{li2026exploring}; training-free remote sensing OVS methods, including SegEarth-OV~\citep{li2025segearth} and SegEarth-OV3~\citep{li2025segearth3}; and the multimodal remote sensing OVS method MM-OVSeg~\citep{wei2026mm}. SegEarth-OV and SegEarth-OV3 are evaluated without additional training, while the remaining models are trained on the corresponding OpenEarth-OVS44 training splits using their official implementations. We report mean Intersection over Union (mIoU) as the primary metric for evaluation of semantic segmentation performance. All experiments are conducted on NVIDIA A100 GPUs with 80 GB memory. 

%We provide more implementation details in the Appendix \ref{sec:implement}.

%----------------------------------------
%-----------------------------------------
\newcommand{\sep}{\hspace{2pt}\rule{0.4pt}{2.1ex}\hspace{2pt}}
\begin{table*}[t]
\centering
\renewcommand{\arraystretch}{1.28}
\setlength{\tabcolsep}{4pt}
\caption{Performance splits for \textit{unseen} and \textit{seen} classes. The table reports mIoU scores for each setting. Settings correspond to:
    \ding{172}: {OpenEarth-OVS44-Optical}; \ding{173}: {OpenEarth-OVS44-Cloudy}; \ding{174}: {OpenEarth-OVS44-Cross}.}
\label{tab:performanceSplit}

\resizebox{\textwidth}{!}{%
\begin{tabular}{l c c c c c c c c c c c c c c c c}
\toprule
\multirow{2}{*}{Method} 
& \multirow{2}{*}{\sep}
& \multirow{2}{*}{Publication} 
& \multirow{2}{*}{\sep}
& \multirow{2}{*}{Backbone} 
& \multirow{2}{*}{\sep}
& \multicolumn{3}{c}{\ding{172}: Clear-sky Condition} 
& \multirow{2}{*}{\sep}
& \multicolumn{3}{c}{\ding{173}: Cloudy Condition} 
& \multirow{2}{*}{\sep}
& \multicolumn{3}{c}{\ding{174}: Cross-region} \\
& & & & & &
Unseen & Seen & Overall
& &
Unseen & Seen & Overall
& &
Unseen & Seen & Overall \\
\midrule
\multicolumn{17}{c}{Zero-shot Testing} \\
\midrule

SegEarth-OV \citep{li2025segearth}
& \sep & CVPR'25 & \sep & ViT-B/16 & \sep
& \textbf{14.4} & \textbf{14.4} & \textbf{14.4}
& \sep
& \textbf{6.6} & \textbf{5.8} & \textbf{6.1}
& \sep
& \textbf{17.0} & \textbf{16.1} & \textbf{16.3} \\

SegEarth-OV3 \citep{li2025segearth3}
& \sep & arXiv'25 & \sep & SAM3 & \sep
& 2.6 & 0.9 & 1.4
& \sep
& 2.2 & 0.6 & 1.0
& \sep
& 4.5 & 0.8 & 1.8 \\

\midrule
\multicolumn{17}{c}{Train on OpenEarth-OVS44 Benchmark} \\
\midrule

CAT-Seg \citep{cho2024catseg}
& \sep & CVPR'24 & \sep & ViT-B/16 & \sep
& 12.4 & \textbf{37.2} & \textbf{29.8}
& \sep
& \textbf{11.6} & 27.8 & \textbf{23.3}
& \sep
& \textbf{12.5} & \textbf{22.3} & \textbf{19.6} \\

EBSeg \citep{shan2024open}
& \sep & CVPR'24 & \sep & ViT-B/16 & \sep
& 3.9 & 36.6 & 26.9
& \sep
& 3.1 & 25.8 & 19.5
& \sep
& 2.8 & 17.7 & 13.6 \\

GSNet \citep{ye2025towards}
& \sep & AAAI'25 & \sep & ViT-B/16 & \sep
& 8.2 & 35.9 & 27.7
& \sep
& 4.0 & 26.0 & 19.8
& \sep
& 9.3 & 18.6 & 16.1 \\

RSKT-Seg \citep{li2026exploring}
& \sep & AAAI'26 & \sep & ViT-B/16 & \sep
& \textbf{12.7} & 37.1 & \textbf{29.8}
& \sep
& 8.0 & \textbf{28.1} & 22.5
& \sep
& 12.1 & 20.0 & 17.9 \\

\midrule

CAT-Seg \citep{cho2024catseg}
& \sep & CVPR'24 & \sep & ViT-L/14 & \sep
& \textbf{17.2} & \textbf{41.6} & \textbf{34.4}
& \sep
& 11.9 & 31.7 & 26.1
& \sep
& \textbf{19.1} & \textbf{29.7} & \textbf{26.8} \\

RSKT-Seg \citep{li2026exploring}
& \sep & AAAI'26 & \sep & ViT-L/14 & \sep
& 14.9 & 39.4 & 32.2
& \sep
& 10.8 & 32.2 & 26.2
& \sep
& 16.1 & 26.5 & 23.7 \\

MM-OVSeg \citep{wei2026mm}
& \sep & CVPR'26 & \sep & ViT-L/14 & \sep
& -- & -- & --
& \sep
& \textbf{13.7} & \textbf{33.4} & \textbf{27.9}
& \sep
& -- & -- & -- \\

\bottomrule
\end{tabular}%
}
\end{table*}
%----------------------------------------
\begin{table}[t]
\centering
\caption{
Effect of taxonomy granularity on the training-free SegEarth-OV model. For the coarse evaluation, the corresponding
fine-grained ground-truth masks are merged into \textit{Road} and
\textit{Water}, and the model is queried using the coarse class names.
}
\label{tab:taxonomy_granularity}
\small
\setlength{\tabcolsep}{5pt}
\renewcommand{\arraystretch}{1.08}
\begin{tabularx}{\linewidth}{@{}lccX@{}}
\toprule
\textbf{Group}
& \textbf{Coarse IoU}
& \textbf{Fine-grained mIoU}
& \textbf{Fine-grained subclasses (IoU)} \\
\midrule

Road
& 91.4
& 6.4
& Roundabout (2.6), Railway (5.3), Bridge (3.4),
  Highway (13.8), Normal Road (13.2), Airport Runway (0.0) \\

Water
& 91.8
& 12.9
& Sea/Lake/Pond (28.7), River (0.6), Swimming Pool (9.5) \\

\bottomrule
\end{tabularx}
\end{table}
%-----------------------------------------
\begin{figure*}[!t]
\centering
\includegraphics[width=\textwidth]{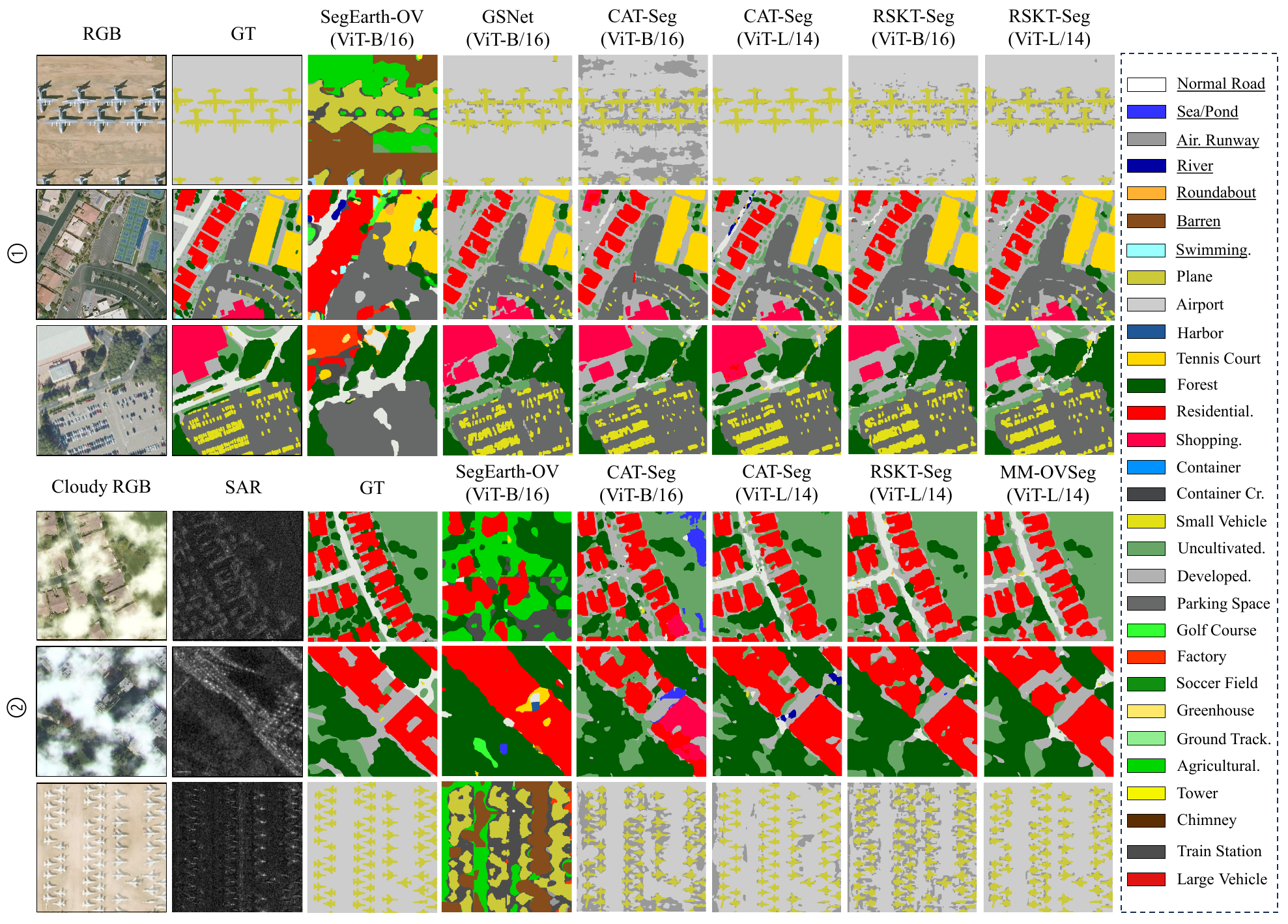}
\caption{{Visualization of OVS results on {OpenEarth-OVS44}.}
Top: results on {OpenEarth-OVS44-Optical} and {OpenEarth-OVS44-Cross}. From left to right: input RGB image, ground truth, and segmentation outputs from SegEarth-OV (ViT-B/16), GSNet (ViT-B/16), CAT-Seg (ViT-B/16), CAT-Seg (ViT-L/14), RSKT-Seg (ViT-B/16), and RSKT-Seg (ViT-L/14).
Bottom: results on {OpenEarth-OVS44-Cloudy}. From left to right: cloudy RGB image, SAR image, ground truth, and segmentation outputs from SegEarth-OV (ViT-B/16), CAT-Seg (ViT-B/16), CAT-Seg (ViT-L/14), RSKT-Seg (ViT-L/14), and MM-OVSeg (ViT-L/14).
In the legend, underlined categories denote \textit{unseen} classes, while the others denote \textit{seen} classes. Settings correspond to:
    \ding{172}: {OpenEarth-OVS44-Optical}; \ding{173}: {OpenEarth-OVS44-Cloudy}; \ding{174}: {OpenEarth-OVS44-Cross}.}
% \caption{Visualization of OVS results on both {OpenEarth-OVS44-Optical} and {OpenEarth-OVS44-Optical}. On the {OpenEarth-OVS44-Optical} benchmark  (upper side), from left to right: input RGB image, ground truth, and segmentation outputs from SegEarth-OV(ViT-B/16), GSNet(ViT-B/16), CAT-Seg(ViT-B/16), CAT-Seg(ViT-L/14), RSKT-Seg(ViT-B/16), and RSKT-Seg(ViT-L/14); On the{OpenEarth-OVS44-Optical} benchmark (lower side), from left to right: input RGB image, input SAR image, ground truth, and segmentation outputs from SegEarth-OV(ViT-B/16), CAT-Seg(ViT-B/16), CAT-Seg(ViT-L/14), RSKT-Seg(ViT-L/14), and MM-OVSeg(ViT-L/14). In the legend, underlined categories represent \textit{unseen} classes and the remaining categories are \textit{seen} classes.}
\label{fig:visual}
\end{figure*}

%----------------------------------------
\begin{figure*}[!t]
\centering
\includegraphics[width=\textwidth]{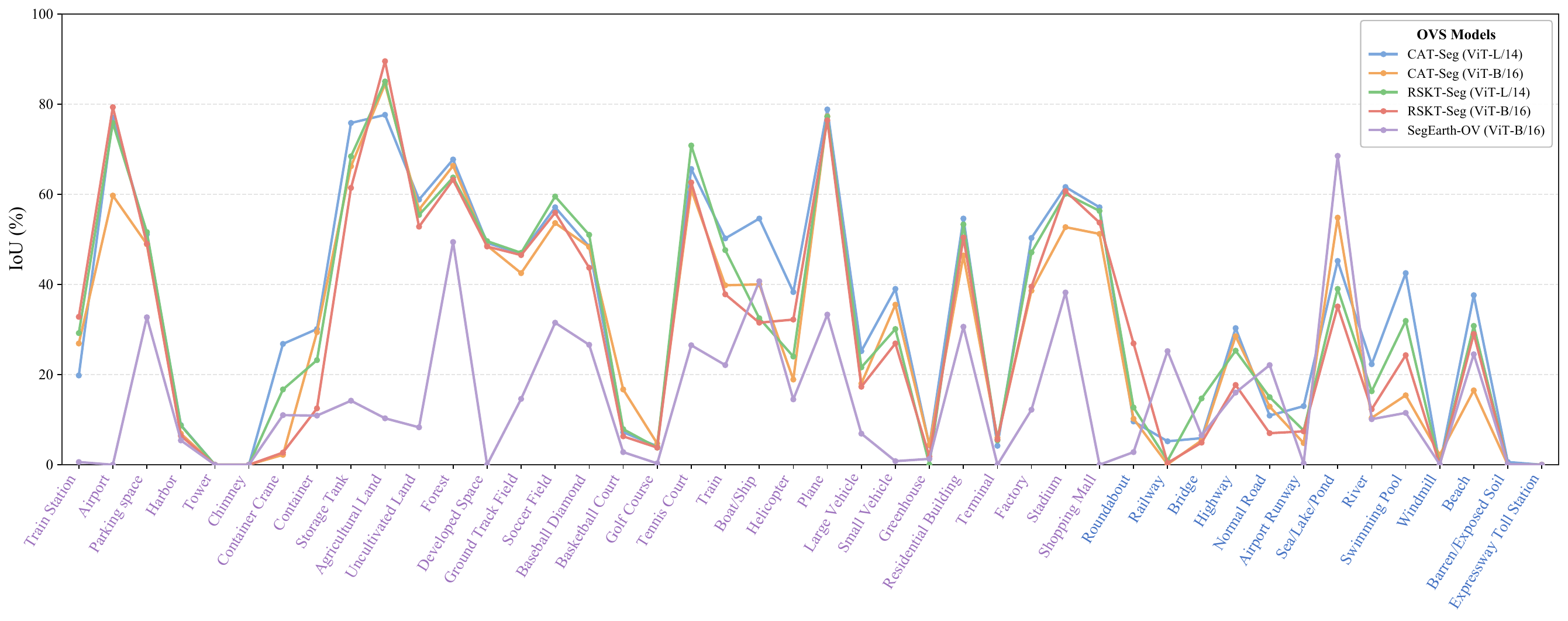}
\caption{Per-class IoU performance of representative OVS models on OpenEarth-OVS44-Optical. \textcolor{perpurple}{Purple} and \textcolor{perblue}{blue} class names represent \textit{seen} and \textit{unseen} classes, respectively.}
\label{fig:Perclass1}
\end{figure*}

% \begin{figure*}[!t]
% \centering
% \includegraphics[width=\textwidth]{Fig-PerClass.pdf}
% \caption{Per-class IoU performance of CAT-Seg (ViT-L/14) on OpenEarth-OVS44-Optical under the nine base categories defined in Table \ref{tab:category_hierarchy}. Each subplot corresponds to one base category and reports the IoU of its refined categories. \textcolor{perpurple}{Purple} bars and \textcolor{perblue}{blue} bars represent \textit{seen} and \textit{unseen} classes, respectively.}
% \label{fig:Perclass1}
% \end{figure*}
%-----------------------------------------

% All training-free OVS models are
\subsection{Main Results}

% We present the quantitative results in Table \ref{tab:performanceSplit} and summarize our key findings as follows:

Table~\ref{tab:performanceSplit} presents the mIoU scores for \textit{seen}, \textit{unseen}, and all categories under the three evaluation settings. We summarize the main observations below.

% \textbf{Current training-free OVS models struggle with fine-grained remote-sensing semantics.} Cutting-edge training-free OVS model achieves overall mIoU scores of 14.4\% on {OpenEarth-OVS44-Optical}, 6.1\% on {OpenEarth-OVS44-Cloudy} and 16.3\% on {OpenEarth-OVS44-Cross}. Notably, replacing the backbone with the more advanced SAM3 \citep{carion2025sam} further degrades performance across benchmarks. This pronounced gap underscores a fundamental limitation of existing training-free OVS methods for fine-grained remote sensing segmentation: simply adopting a stronger frozen backbone is insufficient to improve fine-grained recognition. These results demonstrate the rigor of {OpenEarth-OVS44} and its effectiveness as a benchmark for evaluating fine-grained geospatial understanding.

\textbf{Current training-free OVS models struggle with fine-grained
remote-sensing semantics.}
The strongest training-free baseline, SegEarth-OV, achieves overall mIoU
scores of 14.4 on OpenEarth-OVS44-Optical, 6.1 on
OpenEarth-OVS44-Cloudy, and 16.3 on OpenEarth-OVS44-Cross.
Moreover, its SAM3-based extension, SegEarth-OV3, exhibits even lower
performance across all three evaluation settings, indicating that simply
adopting a stronger frozen vision backbone is insufficient for fine-grained
remote-sensing recognition. To further isolate the effect of semantic granularity, we evaluate SegEarth-OV on the full OpenEarth-OVS44-Optical test set using both coarse- and fine-grained taxonomies, while keeping all other inference settings unchanged. As shown in Table~\ref{tab:taxonomy_granularity}, SegEarth-OV achieves 91.4 and 91.8 IoU on the coarse \textit{Road} and \textit{Water} categories, respectively, but only 6.4 and 12.9 mIoU after they are decomposed into fine-grained subclasses. This substantial gap indicates that coarse taxonomies can mask the inability of existing OVS models to distinguish semantically and functionally distinct subclasses, thereby overstating their fine-grained recognition capability. Together with the training results in Table~\ref{tab:performanceSplit}, these findings highlight the importance of fine-grained taxonomies not only for rigorous evaluation, but also for learning more discriminative remote-sensing semantics.

\textbf{OVS models benefit substantially from training on OpenEarth-OVS44 benchmark.} All training-required models substantially improve performance on \textit{seen} categories compared with training-free baselines. Among unimodal models, CAT-Seg with ViT-L/14 achieves the best overall performance on OpenEarth-OVS44-Optical and OpenEarth-OVS44-Cross, reaching 34.4 and 26.8 overall mIoU, respectively. Among unimodal models, RSKT-Seg with ViT-L/14 achieves the largest \textit{seen}-category improvement on OpenEarth-OVS44-Cloudy, by 26.4 mIoU points. On \textit{unseen} categories, CAT-Seg with ViT-L/14 consistently outperforms the training-free baseline, exceeding it by 2.8 mIoU points on OpenEarth-OVS44-Optical and 2.1 mIoU points on OpenEarth-OVS44-Cross; under the more challenging OpenEarth-OVS44-Cloudy setting, this margin further increases to 5.3 mIoU points. Additionally, CAT-Seg and RSKT-Seg further benefit consistently from scaling the backbone from ViT-B/16 to ViT-L/14, consistent with the visual improvements shown in Figure \ref{fig:visual}. These substantial improvements demonstrate the effectiveness of our dataset for fine-grained segmentation.

% Additionally, RSKT-Seg and CAT-Seg also benefit consistently from scaling the backbone from ViT-B/16 to ViT-L/14. This trend is further supported by the visual comparisons in Figure \ref{fig:visual}, where the ViT-L/14 variants produce clearer boundaries, more stable predictions across diverse weather conditions, and more accurate responses on both \textit{seen} and \textit{unseen} categories. This intuitive improvement demonstrates how increasing model capacity strengthens the fine-grained segmentation capability of training-required OVS models, reinforcing the effectiveness of our dataset for fine-grained segmentation.

% Compared with training-free baselines, all training-required OVS models exhibit substantial improvements on seen categories, as expected. Among the unimodal OVS models, CAT-Seg (ViT-L/14) achieves the largest gains on OpenEarth-OVS44-Optical and OpenEarth-OVS44-Cross, with mIoU improvements of 20.0\% and 10.5\%, respectively, whereas RSKT-Seg (ViT-L/14) yields the largest improvement on OpenEarth-OVS44-Cloudy, with a 26.4\% increase in mIoU. On unseen categories, CAT-Seg (ViT-L/14) consistently outperforms the training-free baseline, exceeding it by 2.8\% on OpenEarth-OVS44-Optical and 2.1\% on OpenEarth-OVS44-Cross in terms of mIoU; under the more challenging OpenEarth-OVS44-Cloudy setting, this margin further increases to 5.3\%. Additionally, CAT-Seg (ViT-L/14) and RSKT-Seg (ViT-L/14) consistently outperform their ViT-B/16 version on both benchmarks due to its increased model capacity.

\textbf{OVS models substantially degrade under low-visibility conditions.} Compared with the clear-sky optical setting, all unimodal models suffer substantial performance degradation under cloudy conditions. Among them, SegEarth-OV suffers the largest performance degradation on \textit{unseen} categories, with a 7.8 drop in mIoU, whereas EBSeg exhibits the largest decline on \textit{seen} categories, with a 10.8 drop in mIoU. In contrast, MM-OVSeg with ViT-L/14 outperforms all unimodal OVS models on OpenEarth-OVS44-Cloudy, with 27.9 overall mIoU. This improvement can be attributed to its effective integration of complementary SAR information, demonstrating the superior robustness and generalization of multimodal fusion under adverse conditions.

% For example, SegEarth-OV with ViT-B/16 drops from 14.4\% to 6.1\% overall mIoU, and CAT-Seg with ViT-L/14 drops from 34.4\% to 26.1\%. MM-OVSeg with ViT-L/14 achieves the state-of-the-art performance on OpenEarth-OVS44-Cloudy, with 27.2\% overall mIoU. 

% Compared with the clear-sky benchmark, all OVS models suffer noticeable performance drops under cloudy conditions. Among them, SegEarth-OV (ViT-B/16) suffers the largest performance degradation on unseen categories, with a 7.8\% drop in mIoU, whereas EBSeg (ViT-B/16) exhibits the largest decline on seen categories, with a 10.8\% drop in mIoU. Despite degraded optical observations, MM-OVSeg (ViT-L/14) still outperforms all unimodal OVS models, with overall mIoU improvements of 1.0\%.

\textbf{Current OVS models still exhibit several limitations.} Figure~\ref{fig:Perclass1} further presents the per-class IoU results of existing models on OpenEarth-OVS44-Optical. Although CAT-Seg with ViT-L/14 outperforms all competing methods by large margins, its performance on \textit{seen} categories remains notably higher than that on \textit{unseen} categories, highlighting the intrinsic difficulty of remote sensing OVS, particularly in preserving robust visual-text alignment. Interestingly, CAT-Seg with ViT-L/14 achieves strong performance on \textit{unseen} water-related categories. This can be attributed to the strong visual-text alignment capability of CLIP \citep{radford2021learning} for such concepts. In contrast, for several rare categories, such as \textit{tower}, \textit{chimney}, and \textit{expressway toll station}, CAT-Seg fails to establish effective visual-text correspondence, resulting in inferior performance. These results further demonstrate the rigor of OpenEarth-OVS44 and reveal two promising directions for improving OVS models: \textbf{1) stronger visual-text alignment for rare categories}:  given the poor performance of existing models on rare classes, incorporating category-specific prior knowledge may help strengthen visual-text correspondence and improve recognition of such categories; and \textbf{2) refined text prompt design}: the text prompts adopted by current OVS models, typically in the form of “a photo of \{class name\},” are overly simplistic and may introduce textual ambiguity, thereby limiting performance on certain categories. Designing more informative and less ambiguous prompts could further improve fine-grained open-vocabulary segmentation.

\section{Conclusion}

In this work, we present {OpenEarth-OVS44}, a fine-grained multimodal remote sensing benchmark for unimodal and multimodal open-vocabulary segmentation. It comprises 290,587 annotated segments from 2,832 RGB images and 1,341 paired RGB-SAR images at sub-meter spatial resolution, covering 44 categories across 40 cities in France, Japan, and the United States. Three evaluation tracks are provided: an RGB-only fine-grained benchmark, a multimodal track with diverse synthetic cloudy conditions for weather-robust evaluation, and a held-out cross-region track. Extensive experiments on seven advanced OVS models highlight both the potential and the inherent limitations of current methods for fine-grained remote sensing segmentation. Additionally, through training five OVS models on {OpenEarth-OVS44}, we demonstrate substantial performance enhancements across all evaluation tracks, with improvements of up to 7.1 mIoU points on \textit{unseen} categories and 27.6 mIoU points on \textit{seen} categories. We expect {OpenEarth-OVS44} and its baselines to provide a strong foundation for future research in fine-grained remote sensing open-vocabulary segmentation.

% \section*{Acknowledgments}
% This was was supported in part by......

%Bibliography
\bibliographystyle{unsrt}  
\bibliography{references}

\end{document}